\documentclass[sigconf]{acmart}

\usepackage{booktabs} 
\usepackage{graphicx}
\usepackage{subfigure}

\setcopyright{rightsretained}

\acmDOI{10.475/123_4}

\acmISBN{123-4567-24-567/08/06}

\acmConference[SIGSPATIAL'17]{ACM SIGSPATIAL conference}{November 2017}{Redondo Beach, California USA} 
\acmYear{2017}
\copyrightyear{2017}

\acmPrice{15.00}

\begin{document}
	\title{Large-Scale Mapping of Human Activity using Geo-Tagged Videos}
	
	\author{Yi Zhu}
	\affiliation{%
		\institution{University of California, Merced}
	}
	\email{yzhu25@ucmerced.edu}
	
	\author{Sen Liu}
	\affiliation{%
		\institution{University of Southern California}
	}
	\email{senliu@usc.edu}
	
	\author{Shawn Newsam}
	\affiliation{%
		\institution{University of California, Merced}
	}
	\email{snewsam@ucmerced.edu}
	
	\renewcommand{\shortauthors}{Yi Zhu, Sen Liu and Shawn Newsam}

	\begin{abstract}
		This paper is the first work to perform spatio-temporal mapping of human activity using the visual content of geo-tagged videos. We utilize a recent deep-learning based video analysis framework, termed hidden two-stream networks, to recognize a range of activities in YouTube videos. This framework is efficient and can run in real time or faster which is important for recognizing events as they occur in streaming video or for reducing latency in analyzing already captured video. This is, in turn, important for using video in smart-city applications. We perform a series of experiments to show our approach is able to accurately map activities both spatially and temporally. We also demonstrate the advantages of using the visual content over the tags/titles.
	\end{abstract}

	
	
	\maketitle
	
	\section{Introduction}
	\label{sec:introduction}
	Mapping human activity on a large scale in real time or near real time is a fundamental yet challenging task in the geographic and social sciences. It is an essential component for making cities smart, particularly with regard to resource allocation, disease control, social interaction, traffic management, etc. 
	
	Researchers have exploited technological advances to map human mobility by using (GPS) trajectory data \cite{Density_Demsar_2010,Trajectory_Qi_2013,LULC_Activity_Li_2017} or mobile phone records \cite{Activity_Aware_Phithakkitnukoon_2010,Mobile_Sagl_2014,Phone_NE_2015,Mobility_Yang_2016,Generative_Yin_2017}. These approaches do not provide information about specific activities though.
	
	Recently, researchers have utilized large-scale geo-referenced multimedia from social networks, like Twitter, Facebook, Instagram and YouTube, to map activity. Our work falls into this category. The public willingly shares their daily activities in real time or near real time by posting words, pictures and videos to these sites. The content of these large collections of multimedia along with the associated metadata such as geo-coordinates, time stamps, tags/titles, popularity rating, etc. thus represents a promising opportunity to map activity.
	
	There has been work on using Twitter to geo-visualize human activity on maps \cite{City_Story_Kling_2012,Twitter_Hawelka_2014}. Since the vast majority of Twitter feeds are only text, and all Tweets are very short, such approaches are not effective for precise activity mapping. Indeed, even though \cite{Twitter_Hawelka_2014} collects millions of geo-tagged Tweets, only global mobility patterns are detected and mapped.
	
	There has also been work on using geo-tagged images to analyze human activity \cite{Flickr_Kisilevich_2010,VGI_Sun_2013,Magnetism_Paldino_2015,Emotions_Hauthal_2016,Travel_Yuan_2016}. While this is closer to our work, it is still limited since most activities have a temporal component which is not captured by an image.
	
	We therefore propose using geo-tagged videos to map human activity. We consider both the appearance and temporal (dynamic) aspects of the videos. This allows more effective activity detection than using tags/titles or the visual content of images.
	
	Performing activity recognition in video is a challenging problem. Video data is large which makes real-time or near real-time analysis difficult. And, video data is very complex. Activity recognition is a high-level vision task which is very difficult to do using the pixel information alone.
	
	Fortunately, the field of computer vision has made great progress recently in high-level video understanding thanks to deep learning. Large-scale labeled video datasets \cite{ucf101,activityNet,KarpathyCVPR14,YouTube_8M_2016} have been created, allowing deep convolutional neural networks (CNN) to be trained and achieve impressive performance on activity recognition. We take advantage of this recent progress to perform, for the first time, spatio-temporal mapping of human activity using geo-referenced videos.
	
	This paper bridges activity recognition in video with geographic knowledge discovery. The salient aspects of the work include:
	
	\begin{itemize}
		\item Our work is the first to perform spatio-temporal mapping of human activity using the visual content of geo-tagged videos.
		\item We utilize a recent, efficient video analysis framework termed hidden two-stream networks. The framework performs activity recognition at $130$ frames per second (fps) which allows it to run in real time. Efficiency is important for analyzing live video streams and reducing latency in analyzing recently captured video.
		\item The video analysis framework is effective, achieving 90 percent accuracy on a 10 class activity classification problem.
		\item We show our approach is able to spatially map a diverse set of sports activities.
		\item We show our approach is able to detect the impact of weather, such as temperature and precipitation, on activity.
		\item We show our approach is able to spatio-temporally map specific events such as parades and street fights.
		\item We show that using video content is more accurate than using tags/titles.
		\item Our framework is flexible. It could easily be adapted to use geo-referenced videos to map a range of activities that are important for smart cities such as monitoring public safety, monitoring traffic, monitoring public health, etc. Further, the framework is robust enough that it does not require traditional surveillance videos but can use crowd-sourced, YouTube-like videos captured by mobile devices.
	\end{itemize}

	\section{Related Work}
	\label{sec:related_work}
	Our work is related to several lines of research.
	
	\noindent \textbf{Large-Scale Geo-Tagged Multimedia}
	The exponential growth of publicly available geo-referenced multimedia has created a range of interesting opportunities to learn about our world. At the intersection of geographic information science and computer vision, large collections of geotagged photos/videos have been used to map world phenomenon \cite{CrandallMapWorld09}, classify land use \cite{landuse_yi_sigspatial_2015}, perform geo-location \cite{Haysim2gps08,geolocalization_Gupta_2016}, model landmarks \cite{3dModelSnavely08}, conduct urban planning \cite{Magnetism_Paldino_2015}, and detect sentiment hotspots \cite{emotion_yi_sigspatial_2016}.
	
	Although such open-access multimedia represents a wealth of information, analyzing it is challenging due to how noisy and diverse it is. Challenges specific to using this data for geographic discovery include inaccurate location information, uneven spatial distribution and varying photographer intent. We are mindful of these challenges and recognize they likely temper our results.
	
	Our work is novel in that it uses a large collection of geo-tagged videos to map human activity as conveyed through the videos that ordinary people take. We specifically focus on spatial and spatio-temporal activity analysis in an urban area.
	
	\noindent \textbf{Visual Geo-localization}
	Geo-localization is the problem of determining where something is. There exists an extensive body of literature on the large-scale visual geo-localization of images. Video geo-localization by comparison is relatively less studied \cite{Overhead_Hammoud_2013,Structure_Bodensteiner_2015,event_More_2016,geolocalization_Gupta_2016}. 
	Note that our goal is not to perform geo-localization. Our videos are already geo-tagged. Our goal is to perform geographic knowledge discovery by analyzing the geo-tagged videos.
	
	\noindent \textbf{Video Activity Recognition}
	The field of human action recognition in video has evolved significantly over the past few years. Traditional handcrafted features such as Improved Dense Trajectories (IDT) \cite{idtfWang2013} dominated the field of video analysis for many years. However, despite their excellent performance, IDT and its improvements \cite{peng2014action,MIFS2015,dovf_lan_2017} are too computationally restrictive to be used for real time applications. CNNs \cite{KarpathyCVPR14,c3d2015}, which are often several orders of magnitude faster than IDTs, initially performed much worse than IDTs. This inferior performance was due mostly to the difficulty appearance-based CNNs have in capturing the movement between frames. Subsequent two-stream CNNs \cite{twostream2014, wanggoodpractice2015} addressed this problem by pre-computing optical flow using traditional optical flow estimation methods \cite{tvl1realTime} and training a separate CNN to encode the movement captured by the optical flow. This additional stream (also known as the temporal stream) significantly improved the accuracy of CNNs and finally allowed them to outperform IDTs on several benchmark action recognition datasets \cite{wanggoodpractice2015,convTwoStream2016,depth2action,TSN2016,res_two_stream_nips16,diba_tle_2016}. These accuracy improvements indicate the importance of temporal motion information for action recognition. Pre-computing optical flow is computational and storage intensive and prevents traditional two-stream networks from running in real time.
	
	In this work, we utilize the recent hidden two-stream networks \cite{hidden_zhu_17} for activity recognition. Our framework is extremely efficient yet maintains competitive accuracy with slower approaches which cannot operate in real time. We compare it for activity recognition with another state-of-the-art real-time activity model named C3D \cite{c3d2015}. The results show the superiority of our method. 
	
	\section{Methodology}
	\label{sec:methodology}
	In this section, we first formulate our problem in Section \ref{sec:problem_formulation}. We then describe our approach towards recognizing human activities in videos in Section \ref{sec:hidden_two_stream_networks}. 
	
	\subsection{Problem Formulation}
	\label{sec:problem_formulation}
	Recall from Section \ref{sec:introduction} that work exists on using volunteered geographic information such as Twitter/Facebook texts and Instagram/Flickr images to perform geographic knowledge discovery. There is little work on using geo-tagged videos and, that which does, does not exploit the visual content. The challenge is in developing an effective and efficient video analysis framework. In this paper, we utilize  hidden two-stream networks \cite{hidden_zhu_17} to overcome these challenges. The ``hidden'' part of the model addresses the efficiency while the ``two-stream'' aspect effectively handles both static appearance and dynamic motion. Details about the framework are provided in Section \ref{sec:hidden_two_stream_networks}.
	
	The overarching goal of our work is to show that geo-referenced videos, such as at YouTube, can be used to spatio-temporally map human activity on a large scale. We select $8$ popular sports, \textbf{baseball, basketball, football, golf, racquetball, soccer, swimming and tennis}, as common human activities to map. We also include the class \textbf{parade} to demonstrate how our approach can trace an event and the class \textbf{street fight} to show direct application to public safety. We thus consider $10$ human activities in total but this could easily be extended to others. The fundamental technical problem we now face is human activity recognition in video. The next few sections describe our solution to this problem.
	
	\begin{figure}[t]
		\centering
		\includegraphics[width=1.0\linewidth,trim=0 0 0 0,clip]{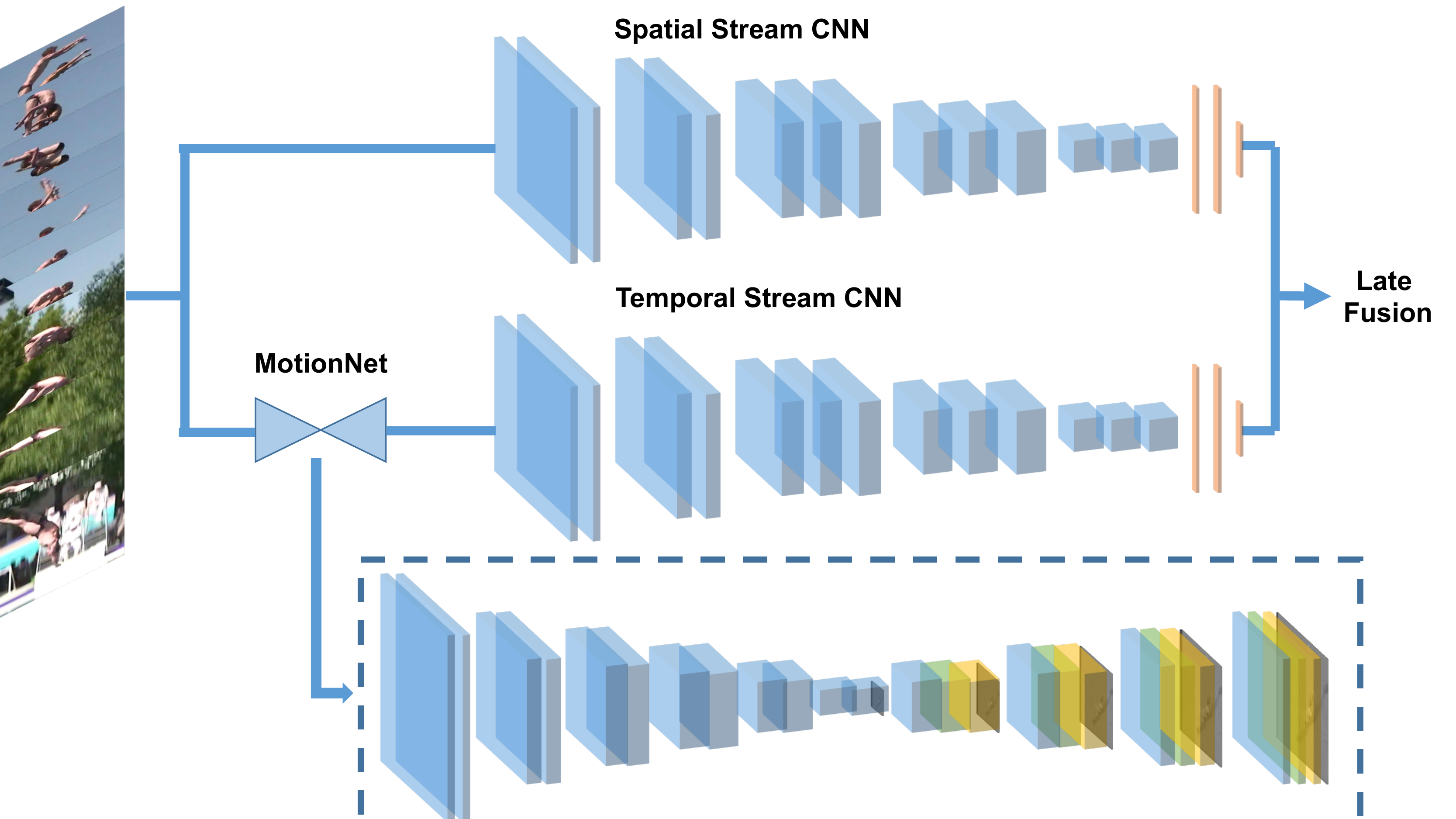}
		\caption{Illustration of the hidden two-stream networks that performs activity recognition using the visual content of a video. The MotionNet CNN takes consecutive video frames as input and outputs the estimated optical flow. The temporal stream CNN then uses this flow to predict activity labels. Late fusion is performed to combine the stacked temporal stream with a spatial stream. Significantly, both streams are end-to-end trainable.}
		\label{fig:framework_hidden}
	\end{figure}
	
	\subsection{Hidden Two-Stream Networks}
	\label{sec:hidden_two_stream_networks}
	In this section, we describe the hidden two-stream networks \cite{hidden_zhu_17} we use for activity recognition in detail. We first recall the baseline two-stream networks \cite{twostream2014} and state its limitation for real world applications in Section \ref{sec:two_stream_networks}. We then introduce the unsupervised network MotionNet for optical flow estimation in Section \ref{sec:motionnet}. In Section \ref{sec:stacked_temporal_stream}, we stack the temporal stream network on top of MotionNet to allow end-to-end training for action recognition. Finally, we combine the stacked temporal stream CNN with the standard spatial stream CNN using late fusion to further improve the activity recognition accuracy.
	
	\subsubsection{\textbf{Two-Stream Networks}}
	\label{sec:two_stream_networks}
	
	As mentioned in Section \ref{sec:related_work}, state-of-the-art activity recognition approaches \cite{wanggoodpractice2015,convTwoStream2016,depth2action,TSN2016,res_two_stream_nips16,diba_tle_2016} build upon the popular two-stream networks framework \cite{twostream2014}. This framework includes a spatial stream CNN which takes RGB video frames (images) as input, and a temporal stream CNN which takes optical flow images as input. This allows the network to model both static appearance and dynamic motion.
	
	However, computing optical flow is significantly more expensive than applying the CNNs. For example, it takes $0.065s$ to estimate a single optical flow image from consecutive $320\times240$ pixel video frames using a GPU accelerated TV-L1 \cite{tvl1realTime} algorithm. This is prohibitively slow for training a temporal stream CNN on-the-fly. The optical flow images also need to be cached before input to the temporal CNN which incurs significant storage costs. The need to pre-compute optical flow is the major speed and storage bottleneck that prevents traditional two-stream approaches from being used in real time applications.
	
	\begin{table}[t]
		\begin{center}
			\resizebox{1.0\columnwidth}{!}{%
				\begin{tabular}{ | c | c | c | c | c | c |  c |}
					\hline
					Name        	&    Kernel & Str & Ch I/O    &    In Res & Out Res   & Input      \\
					\hline		
					conv1 & $3\times3$ & $1$ & $33/64$ & $224\times224$ & $224\times224$ & Frames \\
					conv1$\_$1 & $3\times3$ & $1$ & $64/64$ & $224\times224$ & $224\times224$ & conv1 \\
					conv2 & $3\times3$ & $2$ & $64/128$ & $224\times224$ & $112\times112$ & conv1$\_$1 \\
					conv2$\_$1 & $3\times3$ & $1$ & $128/128$ & $112\times112$ & $112\times112$ & conv2 \\
					conv3 & $3\times3$ & $2$ & $128/256$ & $112\times112$ & $56\times56$ & conv2$\_$1 \\
					conv3$\_$1 & $3\times3$ & $1$ & $256/256$ & $56\times56$ & $56\times56$ & conv3 \\
					conv4 & $3\times3$ & $2$ & $256/512$ & $56\times56$ & $28\times28$ & conv3$\_$1 \\
					conv4$\_$1 & $3\times3$ & $1$ & $512/512$ & $28\times28$ & $28\times28$ & conv4 \\
					conv5 & $3\times3$ & $2$ & $512/512$ & $28\times28$ & $14\times14$ & conv4$\_$1 \\
					conv5$\_$1 & $3\times3$ & $1$ & $512/512$ & $14\times14$ & $14\times14$ & conv5 \\
					conv6 & $3\times3$ & $2$ & $512/1024$ & $14\times14$ & $7\times7$ & conv5$\_$1 \\
					conv6$\_$1 & $3\times3$ & $1$ & $1024/1024$ & $7\times7$ & $7\times7$ & conv6 \\
					flow6 (loss6) & $3\times3$ & $1$ & $1024/20$ & $7\times7$ & $7\times7$ & conv6$\_$1 \\
					deconv5 & $4\times4$ & $2$ & $1024/512$ & $7\times7$ & $14\times14$ & conv6$\_$1 \\
					xconv5  & $3\times3$ & $1$ & $1044/512$ & $14\times14$ & $14\times14$ & deconv5+flow6+conv5$\_$1 \\
					flow5 (loss5) & $3\times3$ & $1$ & $512/20$ & $14\times14$ & $14\times14$ & xconv5 \\
					deconv4 & $4\times4$ & $2$ & $512/256$ & $14\times14$ & $28\times28$ & xconv5 \\
					xconv4  & $3\times3$ & $1$ & $788/256$ & $28\times28$ & $28\times28$ & deconv4+flow5+conv4$\_$1 \\
					flow4 (loss4) & $3\times3$ & $1$ & $256/20$ & $28\times28$ & $28\times28$ & xconv4 \\
					deconv3 & $4\times4$ & $2$ & $256/128$ & $28\times28$ & $56\times56$ & xconv4 \\
					xconv3  & $3\times3$ & $1$ & $404/128$ & $56\times56$ & $56\times56$ & deconv3+flow4+conv3$\_$1 \\
					flow3 (loss3) & $3\times3$ & $1$ & $128/20$ & $56\times56$ & $56\times56$ & xconv3 \\
					deconv2 & $4\times4$ & $2$ & $128/64$ & $56\times56$ & $112\times112$ & xconv3 \\
					xconv2  & $3\times3$ & $1$ & $212/64$ & $112\times112$ & $112\times112$ & deconv2+flow3+conv2$\_$1 \\
					flow2 (loss2) & $3\times3$ & $1$ & $64/20$ & $112\times112$ & $112\times112$ & xconv2 \\
					\hline
					flow2$\_$norm & $3\times3$ & $1$ & $20/20$ & $112\times112$ & $224\times224$ & flow2 \\
					conv1$\_$1$\_$vgg & $3\times3$ & $1$ & $20/64$ & $224\times224$ & $224\times224$ & flow2$\_$norm \\
					conv1$\_$2$\_$vgg & $3\times3$ & $1$ & $64/64$ & $224\times224$ & $224\times224$ & conv1$\_$1$\_$vgg \\
					pool1$\_$vgg & $2\times2$ & $2$ & $64/64$ & $224\times224$ & $112\times112$ & conv1$\_$2$\_$vgg \\
					conv2$\_$1$\_$vgg & $3\times3$ & $1$ & $64/128$ & $112\times112$ & $112\times112$ & pool1$\_$vgg \\
					conv2$\_$2$\_$vgg & $3\times3$ & $1$ & $128/128$ & $112\times112$ & $112\times112$ & conv2$\_$1$\_$vgg \\
					pool2$\_$vgg & $2\times2$ & $2$ & $128/128$ & $112\times112$ & $56\times56$ & conv2$\_$2$\_$vgg \\
					conv3$\_$1$\_$vgg & $3\times3$ & $1$ & $128/256$ & $56\times56$ & $56\times56$ & pool2$\_$vgg \\
					conv3$\_$2$\_$vgg & $3\times3$ & $1$ & $256/256$ & $56\times56$ & $56\times56$ & conv3$\_$1$\_$vgg \\
					conv3$\_$3$\_$vgg & $3\times3$ & $1$ & $256/256$ & $56\times56$ & $56\times56$ & conv3$\_$2$\_$vgg \\
					pool3$\_$vgg & $2\times2$ & $2$ & $256/256$ & $56\times56$ & $28\times28$ & conv3$\_$3$\_$vgg \\
					conv4$\_$1$\_$vgg & $3\times3$ & $1$ & $256/512$ & $28\times28$ & $28\times28$ & pool3$\_$vgg \\
					conv4$\_$2$\_$vgg & $3\times3$ & $1$ & $512/512$ & $28\times28$ & $28\times28$ & conv4$\_$1$\_$vgg \\
					conv4$\_$3$\_$vgg & $3\times3$ & $1$ & $512/512$ & $28\times28$ & $28\times28$ & conv4$\_$2$\_$vgg \\
					pool4$\_$vgg & $2\times2$ & $2$ & $512/512$ & $28\times28$ & $14\times14$ & conv4$\_$3$\_$vgg \\
					conv5$\_$1$\_$vgg & $3\times3$ & $1$ & $512/512$ & $14\times14$ & $14\times14$ & pool4$\_$vgg \\
					conv5$\_$2$\_$vgg & $3\times3$ & $1$ & $512/512$ & $14\times14$ & $14\times14$ & conv5$\_$1$\_$vgg \\
					conv5$\_$3$\_$vgg & $3\times3$ & $1$ & $512/512$ & $14\times14$ & $14\times14$ & conv5$\_$2$\_$vgg \\
					pool5$\_$vgg & $2\times2$ & $2$ & $512/512$ & $14\times14$ & $7\times7$ & conv5$\_$3$\_$vgg \\
					fc6$\_$vgg & $3\times3$ & $1$ & $512/4096$ & $7\times7$ & $1\times1$ & pool5$\_$vgg \\
					fc7$\_$vgg & $3\times3$ & $1$ & $4096/4096$ & $1\times1$ & $1\times1$ & fc6$\_$vgg \\
					fc8$\_$vgg (action$\_$loss) & $3\times3$ & $1$ & $4096/M$ & $1\times1$ & $1\times1$ & fc7$\_$vgg \\
					\hline
				\end{tabular}
			} 
			\vspace{2ex}
			\caption{Stacked temporal stream CNN architecture. Top: MotionNet CNN. Bottom: traditional temporal stream CNN. M is the number of activity classes. Str: stride. Ch I/O: number of channels in the input/output feature maps. In/Out Res: input/output resolution.  \label{tab:stacked_model}}
			\vspace{-6ex}
		\end{center}
	\end{table} 
	
	\subsubsection{\textbf{MotionNet}}
	\label{sec:motionnet}
	
	In order to achieve real time activity recognition, we use the MotionNet \cite{hidden_zhu_17} CNN instead of slower, handcrafted methods to compute optical flow. The key to using a CNN is to pose optical flow computation as a learning problem. MotionNet treats motion estimation as an image reconstruction problem \cite{jasonUnsup2016,guided_flow_17,densenet_denseflow_icip17} where we seek to learn the optimal optical flow that allows the current video frame to be constructed from the previous one. Formally, given a pair of adjacent video frames $I_{1}$ and $I_{2}$ as input, MotionNet generates a motion field $V$. $V$ and $I_{2}$ are then used to produce the estimate $I_{1}^{\prime}$ using inverse warping, i.e., $I_{1}^{\prime} = \mathcal{T}[I_{2}, V]$, where $\mathcal{T}$ is the inverse warping function. The goal is to minimize the photometric (pixelwise) error between $I_{1}$ and $I_{1}^{\prime}$. 
	
	The architectural details of MotionNet can be seen in Table \ref{tab:stacked_model}. The top section of the table corresponds to MotionNet and the bottom section is the traditional temporal stream CNN upon which MotionNet is stacked. Training MotionNet to learn optimal optical flow involves minimizing the following three objective functions:
	\begin{itemize}
		\item A standard pixelwise reconstruction error function
		\begin{equation}
			L_{\text{pixel}} = \frac{1}{N} \sum_{i, j}^{N} \rho ( I_{1}(i, j) - I_{2}(i+V_{i,j}^{x}, j+V_{i,j}^{y}) )
			\label{eq:pixel_loss}
		\end{equation}
		where $i$ and $j$ are the frame numbers and $V^x$ and $V^y$ are the estimated optical flows in the horizontal and vertical directions. The inverse warping is performed using a spatial transformer module \cite{stn_nips15}. We use a robust convex error function, the generalized Charbonnier penalty $\rho(x) = (x^{2} + \epsilon^{2})^{\alpha}$, to reduce the influence of outliers. 
		$N$ denotes the total number of pixels.
		
		\item A smoothness loss to address the ambiguity of estimating motion in non-textured regions (the aperture problem)
		\begin{equation}
			L_{\text{smooth}} = \rho (\nabla V_{x}^{x} ) + \rho ( \nabla V_{y}^{x}) + \rho ( \nabla V_{x}^{y}) + \rho ( \nabla V_{y}^{y})
			\label{eq:smoothness_loss}
		\end{equation}
		where $\nabla V_{x}^{x}$ and $\nabla V_{y}^{x}$ are the gradients of the estimated flow field $V^{x}$ in the horizontal and vertical directions. Similarly, $\nabla V_{x}^{y}$ and $\nabla V_{y}^{y}$ are the gradients of $V^{y}$. A generalized Charbonnier penalty $\rho(x)$ is also used.
		
		\item A structural similarity (SSIM) loss \cite{SSIM_2004} that helps MotionNet learn the structures of frames. It is calculated as
		
		\begin{equation}
			L_{\text{ssim}} = \frac{1}{N} \sum ( 1 - \text{SSIM} (I_{1}, I_{1}^{\prime}) )
			\label{eq:ssim_loss}
		\end{equation}
		
		where SSIM($\cdot$) is a standard structural similarity function. This forces MotionNet to produce flow fields with clear motion boundaries.
	\end{itemize}
	
	The overall loss is a weighted sum of the pixelwise reconstruction loss, the pixelwise smoothness loss and the region-based SSIM loss 
	\begin{equation}
		L = \lambda_{1} \cdot L_{\text{pixel}} + \lambda_{2} \cdot L_{\text{smooth}}  + \lambda_{3} \cdot L_{\text{ssim}}
		\label{eq:total_unsup_loss}
	\end{equation}
	where $\lambda_{1}$, $\lambda_{2}$ and $\lambda_{3}$ weight the relative importance of the different metrics during training. $\lambda_{1}$ and $\lambda_{3}$ are set to $1$. $\lambda_{2}$ is set as suggested in \cite{flownet}.
	
	\subsubsection{\textbf{Stacked Temporal Stream}}
	\label{sec:stacked_temporal_stream}
	Since MotionNet and the temporal stream are both CNNs, they can be stacked on top of each other and trained in an end-to-end manner.
	%
	%
	%
	%
	The stacked temporal stream CNN is then combined with a standard spatial stream CNN as shown in Figure \ref{fig:framework_hidden}. Following previous literature, the two streams are combined through weighted average late fusion using a spatial to temporal ratio of $1$:$1.5$ as in \cite{wanggoodpractice2015,TSN2016,diba_tle_2016}.
	
	\begin{figure*}[t]
		\centering
		\subfigure[Baseball]{\includegraphics[width=0.32\linewidth,trim=0 25 0 0,clip]{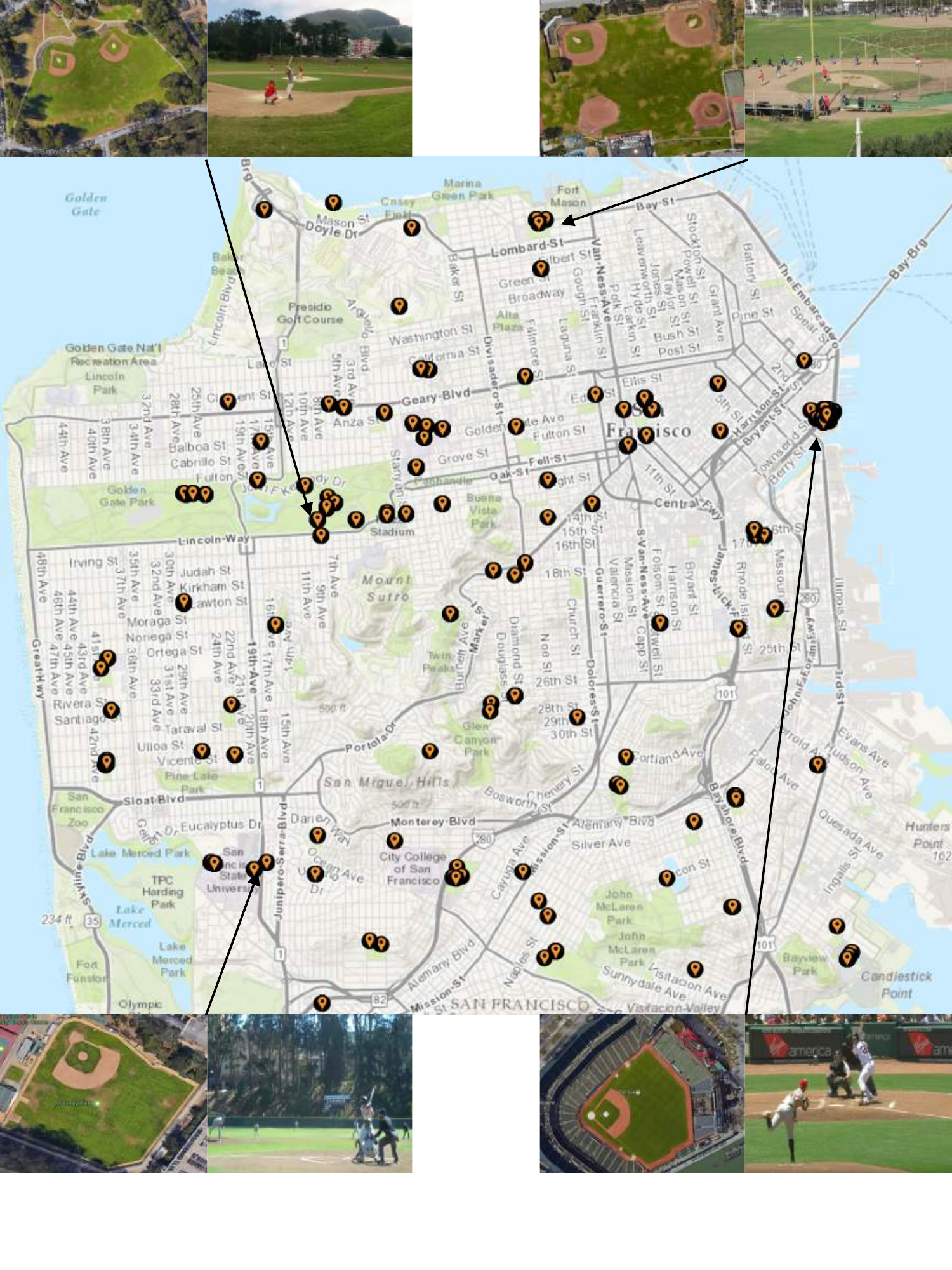}\label{fig:baseball}}\hspace{1pt}
		\subfigure[Basketball]{\includegraphics[width=0.32\linewidth,trim=0 25 0 0,clip]{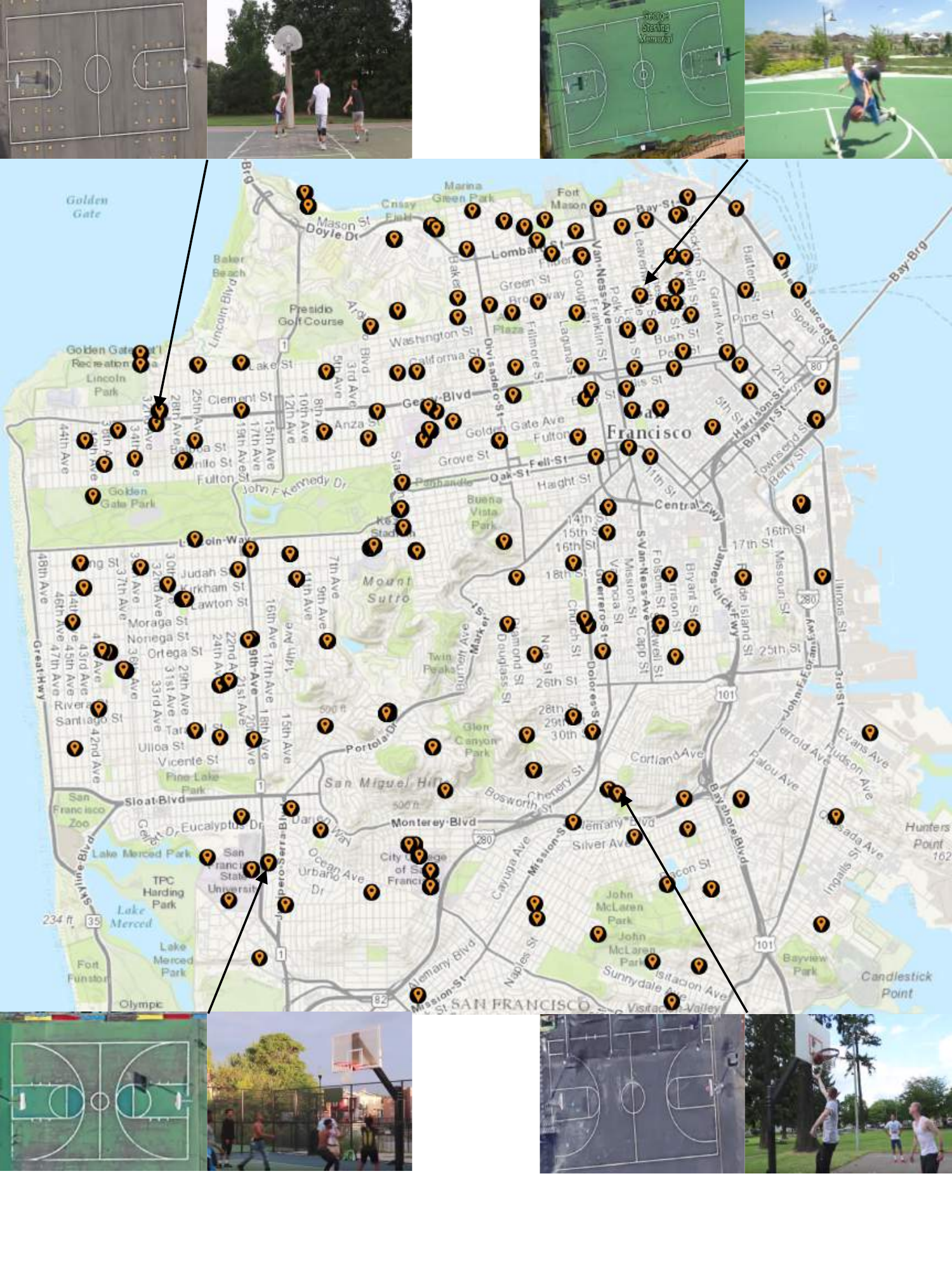}\label{fig:basketball}}\hspace{1pt}
		\subfigure[Football]{\includegraphics[width=0.32\linewidth,trim=0 25 0 0,clip]{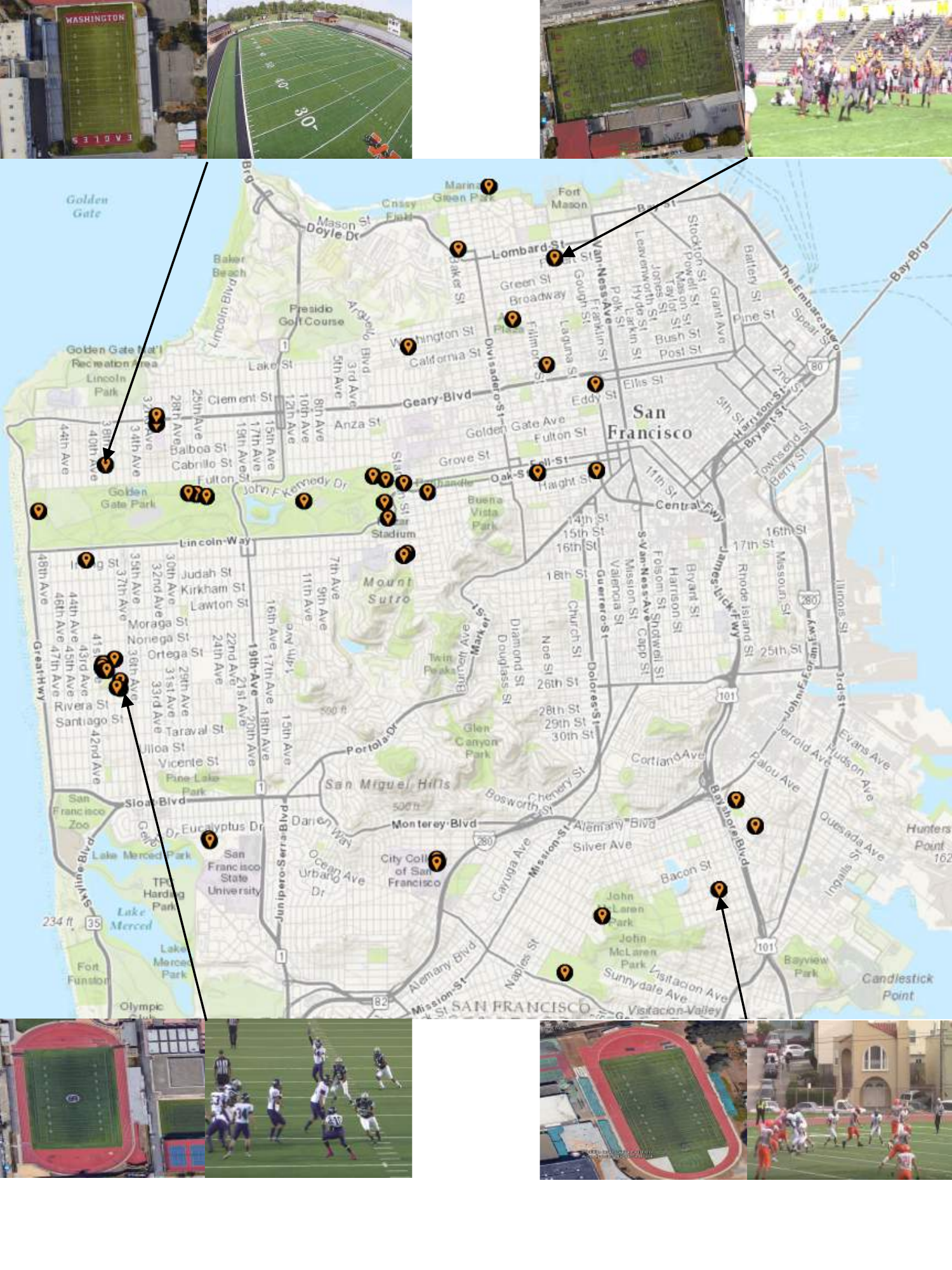}\label{fig:football}}\hspace{1pt}
		\subfigure[Golf]{\includegraphics[width=0.32\linewidth,trim=0 25 0 0,clip]{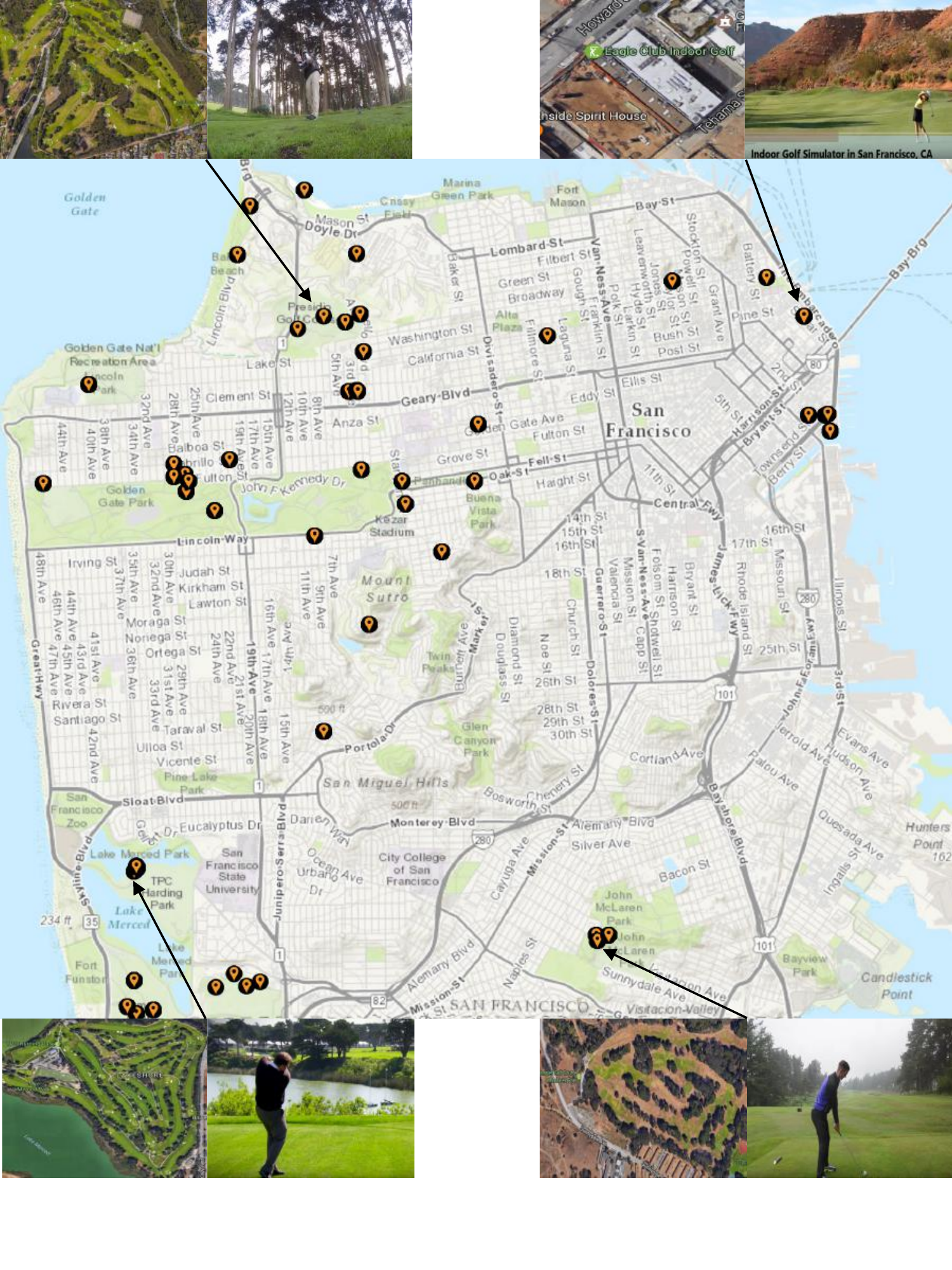}\label{fig:golf}}\hspace{1pt}
		\subfigure[Soccer]{\includegraphics[width=0.32\linewidth,trim=0 25 0 0,clip]{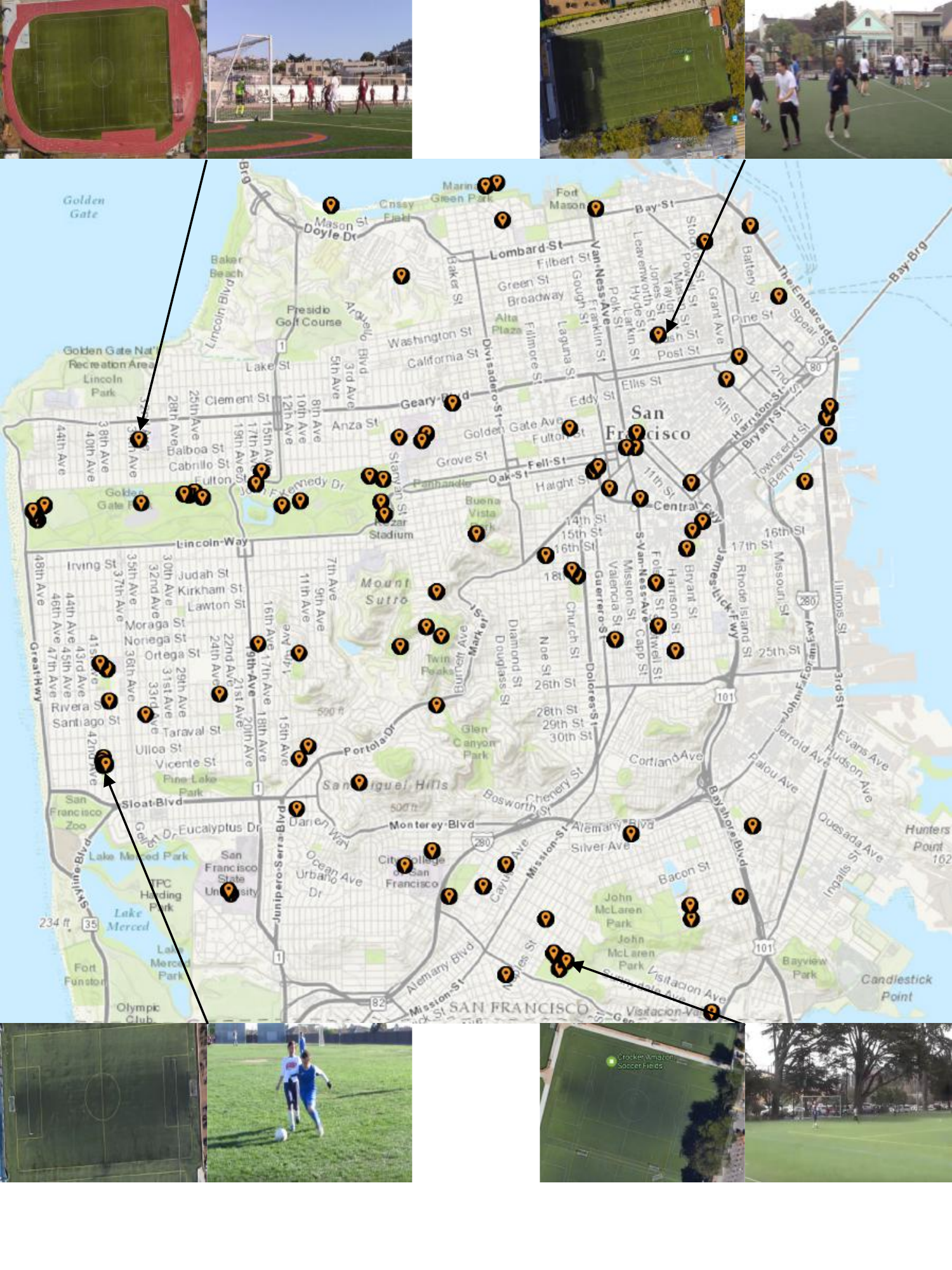}\label{fig:soccer}}\hspace{1pt}
		\subfigure[Tennis]{\includegraphics[width=0.32\linewidth,trim=0 25 0 0,clip]{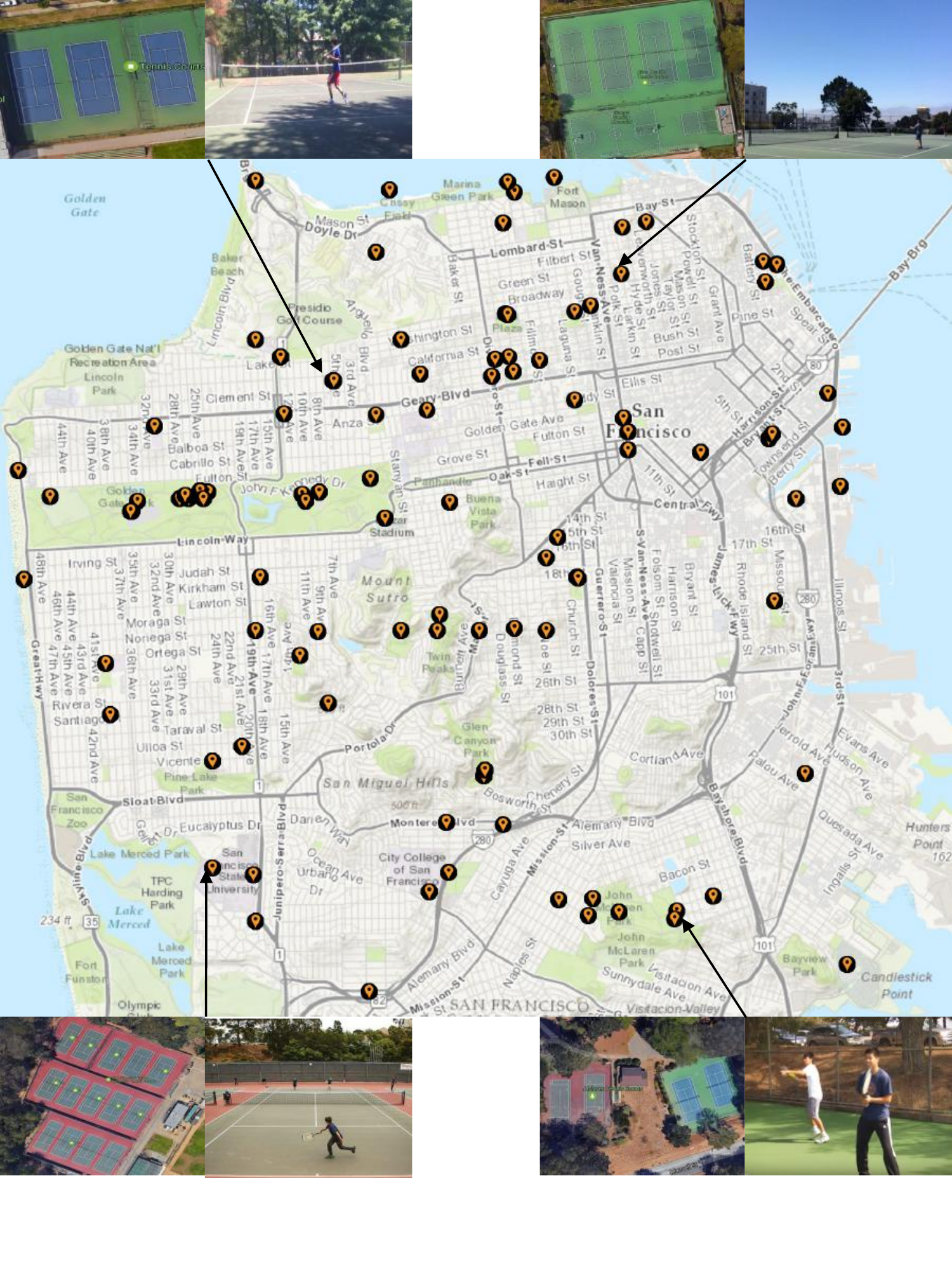}\label{fig:tennis}}\hspace{1pt}
		\vspace{-1ex}
		\caption{Spatial mapping of popular sport in the city of San Francisco for 2016. (a) Baseball; (b) Basketball; (c) Football; (d) Golf; (e) Soccer; and (f) Tennis. Four detections are shown for each sport. We show a sample frame from the video that resulted in the detection as well as a satellite image of the location of the video. This figure is best viewed in color.}
		\label{fig:spatialActivityMap}
	\end{figure*}
	
	\begin{table}[t]
		\begin{center}
			\resizebox{0.8\columnwidth}{!}{%
				\begin{tabular}{ | c | c | c|}
					\hline
					Method										&    Accuracy (\%)  &    fps \\
					\hline
					C3D    					                    &   $84.57$ 	&    $390.7$\\
					Hidden Two-Stream CNNs     					&   $90.94$ 	&    $130.56$\\	
					\hline
				\end{tabular}
			} 
			\vspace{2ex}
			\caption{Comparison of accuracy and efficiency on the 10 class validation dataset. fps stands for frame per second. \label{tab:comparison}}
			\vspace{-4ex}
		\end{center}
	\end{table} 
	
	\begin{table}[t]
		\begin{center}
			\resizebox{0.6\columnwidth}{!}{%
				\begin{tabular}{ | c | c | c|}
					\hline
					Method										&    C3D (\%)   &    Hidden (\%) \\
					\hline
					Baseball    					           &   $86.57$ 	&    $90.29$\\
					Basketball    					           &   $88.62$ 	&    $94.43$\\
					Football    					           &   $85.14$ 	&    $88.63$\\
					Golf    					                &   $76.32$ 	&    $85.91$\\
					Racquetball    					           &   $88.90$ 	&    $92.11$\\
					Soccer    					                &   $83.49$ 	&    $89.43$\\
					Swimming    					           &   $92.27$ 	&    $97.51$\\
					Tennis    					                &   $80.19$ 	&    $84.60$\\
					Parade    					                &   $83.00$ 	&    $90.37$\\
					Street Fight    					           &   $81.20$ 	&    $96.12$\\
					\hline
					Mean Average    					           &   $84.57$ 	&    $90.94$\\
					\hline
				\end{tabular}
			} 
			\vspace{2ex}
			\caption{Per-class accuracy on the validation dataset. \label{tab:break_down}}
			\vspace{-4ex}
		\end{center}
	\end{table} 
	
	\section{Experiments}
	\label{sec:experiments}
	This section first introduces our video dataset. 
	We train and validate our activity recognition model using a large collection of videos with known classes, built from existing activity recognition datasets as well as self-crawled YouTube videos. 
	This section also describes our implementation of the hidden two-stream networks, reports its performance and compares it with another real-time framework called C3D. This section finally presents the results of a number of application scenarios including the spatio-temporal mapping of sports activities, parade events, violence and the effects of weather.
	
	\subsection{Dataset}
	\label{sec:dataset}
	The dataset we use to train and validate our activity recognition model contains $10$ activity classes: baseball, basketball, football, golf, racquetball, soccer, swimming, tennis, parade and street fight. It includes videos from existing activity recognition datasets as well as self-crawled YouTube videos. We only need the activity labels of these videos--we do not need geo-tags. We first leverage existing datasets including Sports-1M \cite{KarpathyCVPR14}, UCF101 \cite{ucf101} and FCVID \cite{FCVID} to create an initial dataset. This initial dataset is too small and unbalanced though for fine-tuning deep CNNs and so we also download YouTube videos\footnote{We limit the duration of downloaded videos to be shorter than ten minutes and to be of high quality.} using the activity labels as keywords. We perform a visual sanity check to remove irrelevant videos from these downloads. Our final dataset contains $10,000$ videos in total, $1,000$ for each activity class. This size is of similar order to the UCF101 \cite{ucf101} and ActivityNet 1.3 \cite{activityNet} datasets which have been shown to be large enough to fine-tune deep networks. We divide this dataset into training and validation components using a split ratio of $0.8$:$0.2$.
	
	To perform our spatio-temporal mapping, we download all geo-tagged YouTube videos using the same keywords within the city of San Francisco for the year $2016$. This results in $265,477$ geo-tagged videos. Note that these videos are disjoint from the ones used to train and validate the activity recognition model above. 
	
	\subsection{Implementation}
	\label{sec:implementation}
	We use the Caffe toolbox \cite{jia2014caffe} to implement the CNNs. All timing results correspond to a workstation with an Intel Core I7 (4.00GHz) and an Nvidia Titan X GPU. 
	
	\noindent \textbf{MotionNet:} 
	MotionNet is trained from scratch using the three unsupervised objectives: the pixelwise reconstruction loss $L_{\text{pixel}}$, the piecewise smoothness loss $L_{\text{smooth}}$ and the region-based SSIM loss $L_{\text{ssim}}$. The generalized Charbonnier parameter $\alpha$ is set to $0.4$ in the pixelwise reconstruction loss, and $0.3$ in the smoothness loss. 
	
	The models are trained using Adam optimization with the default parameter values $\beta_{1}=0.9$ and $\beta_{2}=0.999$. The batch size is $16$. The initial learning rate is set to $3.2\times10^{-5}$ and is divided in half every $100$k iterations. We end our training at $400$k iterations.
	
	\noindent \textbf{Hidden two-stream networks:} 
	The hidden two-stream networks includes the spatial stream and the stacked temporal stream. The MotionNet is pretrained as above. The spatial stream CNN is a VGG16 CNN pretained on the ImageNet challenge \cite{imagenet_cvpr09}, and the stream temporal CNN is initialized with the snapshot provided by Wang \cite{wanggoodpractice2015}. We use stochastic gradient descent to train the networks with a batch size of $128$ and momentum of $0.9$.  We also use horizontal flipping, corner cropping and multi-scale cropping as data augmentation to prevent over fitting.
	
	For the spatial stream CNN, the initial learning rate is set to $0.001$, and is divided by $10$ every $4$K iterations. We stop the training at $10$K iterations. 
	For the stacked temporal stream CNN, we set different initial learning rates for MotionNet and the temporal stream: $10^{-6}$ and $10^{-3}$. We divide the learning rates by $10$ after $5$K and $10$K iterations. The maximum iteration is set to $16$K.
	
	\noindent \textbf{C3D:} 
	C3D \cite{c3d2015} is a generic video analysis framework and consists of 3D convolutions instead of 2D convolutions as in most deep networks. The input to the network is sets of contiguous video frames organized as clips. The model has eight convolutional and five pooling layers which are followed by three fully connected layers. All 3D convolution filters are $3 \times 3 \times 3$ with stride 1, and all 3D poling layers are $2\times2\times2$ with stride 1, except for pool1. In order to preserve temporal information in the earlier layers, pool1 has size $1 \times 2 \times 2$. See \cite{c3d2015} for more details.
	
	C3D is pre-trained on Sports-1M, which has $487$ classes. We fine tune it on our training data. During fine tuning, the network weights are learned using mini-batch (30 video clips) stochastic gradient descent with momentum (set to 0.9). The learning rate is initialized to 0.0001, except for the last fully connected layer with 0.01. Both learning rates decrease to 1/10 of their values whenever the performance saturates, and training is stopped when the learning rate is smaller than $10^{-6}$. Dropout is applied with a ratio of $0.7$ after the first two fully connected layers.
	
	\subsection{Activity Recognition Evaluation}
	\label{sec:activity_recognition_evaluation}
	Table \ref{tab:comparison} compares the accuracy and efficiency of the hidden two-stream networks with the popular C3D network. The hidden two-stream networks achieves just over $90\%$ accuracy on the 10 class validation dataset. It is about $6\%$ more accurate than C3D. C3D is seen to be more efficient but both can run much faster than real time (30 fps). 
	
	Table \ref{tab:break_down} provides the per-class accuracy. Both approaches obtain the highest accuracy on swimming and racquetball. This is likely due to context since swimming is the only water-related sport and racquetball has the largest proportion of indoor scenes. Interestingly, the hidden two-stream networks achieves much higher accuracy than C3D on street fight. This is likely due to the increased capacity of the MotionNet CNN to learn optimal motion representations. Street fight contains significant motion (both subject and camera) which might not be captured by the 3D convolutional filtering performed by C3D.
	
	The remainder of the experiments are performed with the hidden two-stream networks.
	
	\begin{figure}[t]
		\centering
		\includegraphics[width=1.0\linewidth,trim=0 0 0 0,clip]{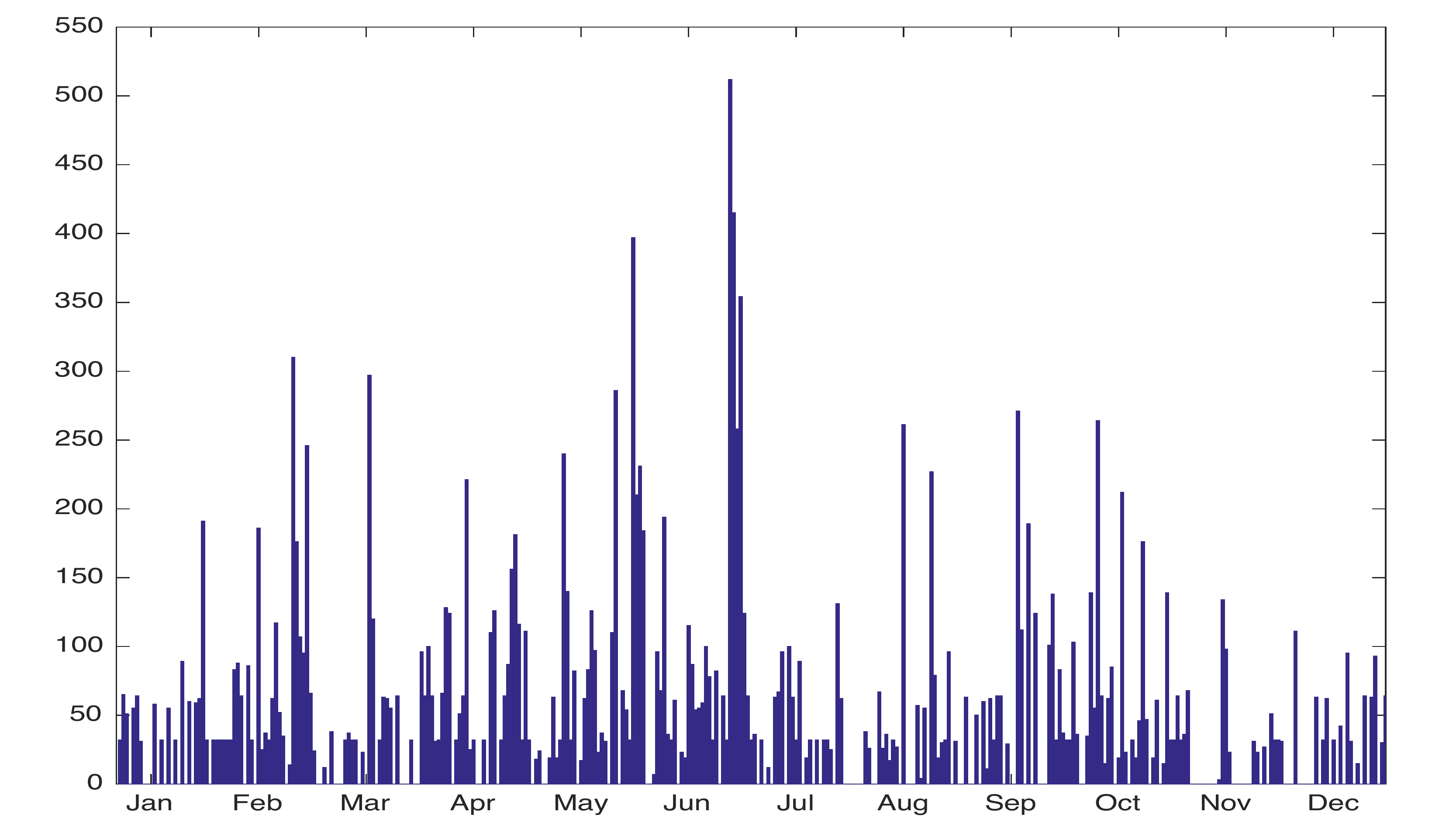}
		\caption{Temporal analysis of user uploaded parade videos in the city of San Francisco in year 2016. The y axis indicates the number of geo-tagged parade videos for each day. The peaks correspond to the major parades.}
		\label{fig:parade_temporal}
	\end{figure}
	
	\begin{figure}[t]
		\centering
		\includegraphics[width=1.0\linewidth,trim=0 0 0 0,clip]{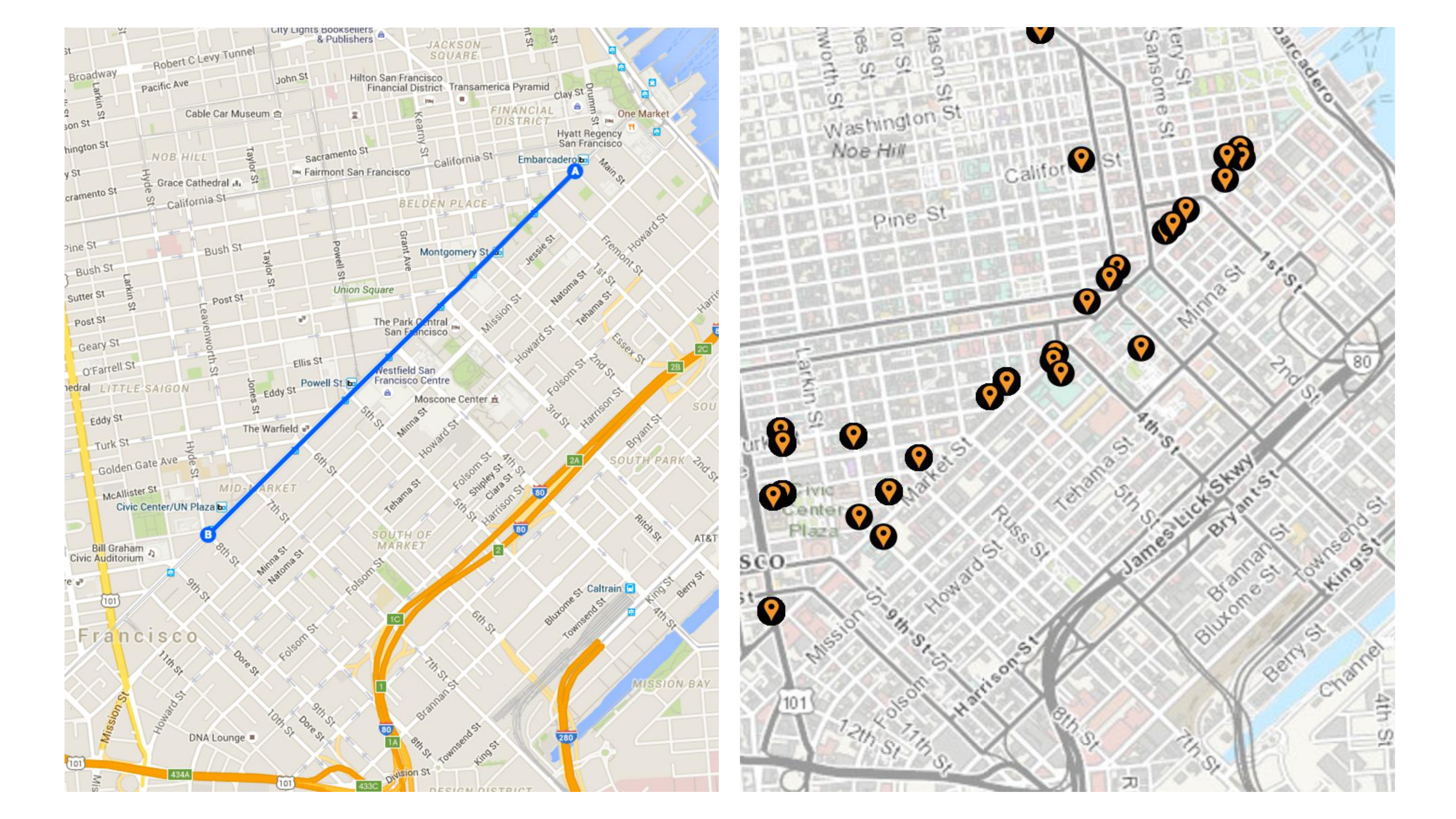}
		\caption{Spatial analysis of the 46th San Francisco Pride parade in 2016. Left: official parade route. Right: map of classified videos. Note the correlation. }
		\label{fig:parade_spatial}
	\end{figure}
	
	\subsection{Spatial Sports Mapping}
	\label{sec:spatial_sports_mapping}
	We now apply our framework to the geo-tagged YouTube videos from San Francisco for 2016. During inference, we sample frames/clips every one second to reduce computational cost.
	Figure \ref{fig:spatialActivityMap} shows the locations of videos classified as the six most popular sports: baseball, basketball, football, golf, soccer and tennis. Also shown are four detections for each sport. We show a sample frame from the video that resulted in the detection as well as a satellite image of the location of the video. These results demonstrate that our approach a) is able to correctly classify the YouTube videos, and b) can use this classification to map where the activities take place.
	
	We make the following four observations based on the results in Figure \ref{fig:spatialActivityMap}.
	
	\noindent \textbf{Observation 1}: Our approach is able to locate sports fields and complexes using the visual content of the geo-tagged videos. Figure \ref{fig:baseball} contains a concentration of points in the area of AT$\&$T park, the home of the SF Giants baseball team. We also locate the San Francisco State University basketball court, George Washington High School football field, TPC Harding Park golf course, Crocker Amazon soccer fields, John McLaren Park tennis courts, etc. 
	
	We are also able to locate where sports are played in a more informally place. For example, there are a lot of videos labeled as basketball in high density residential neighborhoods. We checked these and, indeed, found that they are located in backyards and other places where basketball can be played.
	
	
	\noindent \textbf{Observation 2}: There are relatively few videos classified as golf and football. This makes sense for golf since it requires a large, open area and good weather. The reason for so few football videos is likely that the training videos for this class from the Sports1M dataset are captured during actual games in stadiums with full teams of players wearing helmets, etc. This is different from the more informal types of football activities found in an urban area such as touch football or simply throw and catch. A more representative training dataset is needed for this class.
	
	
	\noindent \textbf{Observation 3}: The video frames and satellite images are in agreement with the predicted sports and their locations. There is, however, one interesting exception (the top right example in Figure \ref{fig:golf}) of a golf video located in downtown San Francisco. Upon further investigation, we found this makes sense since there is an indoor driving range inside the building named Eagle Club indoor golf. The club provides customized scenes and projects virtual driving ranges on large monitors for players to practice. The classified video is an advertisement. This example demonstrates a distinct advantage that ground-level images and videos have over satellite or aerial images--they can be used to perform geographic discovery indoors.
	
	\noindent \textbf{Observation 4}: Our approach is able to use context to detect where a sport is played even if it is not occurring at the time the video was captured or the activity is difficult to discern. For example, in the top left example in Figure \ref{fig:football}, the video snippet is an oblique view of just the football field. And, in the bottom right example in Figure \ref{fig:soccer}, the players are very far from the camera. The ability of our approach to do this can be attributed to the spatial stream CNN's capacity to learn the static appearance of where sports are played.
	
	\subsection{Spatio-Temporal Parade Mapping}
	\label{sec:spatio_temporal_parade_mapping}
	The goal here is to locate specific events, such as a parade, both spatially and temporally. We first detect all parade videos and temporally group them by date. We then map the videos in a group to identify the parade route.
	
	\noindent \textbf{Temporal Analysis}: We detect a total of $15,645$ parade videos in San Francisco in 2016. The daily distribution is shown in Figure \ref{fig:parade_temporal}. The peaks correlate with known parades including the Chinese New Year parade (February 20), the St. Patrick's Day parade (March 12), the Carnaval Grand parade (May 28), the Pride parade (June 25) and the Italian Heritage parade (October 9).
	
	Closer analysis shows that the videos of a parade tend to be uploaded after the event, sometimes days later. This is different from texts or images which tend to be shared during the event. This is likely because video requires better network connectivity. Also, users often first edit their videos before uploading them.
	
	
	\noindent \textbf{Spatial Analysis}: We now map the videos of the most popular parade in San Francisco in 2016 (based on our detections), the 46th Pride parade. As shown in Figure \ref{fig:parade_spatial}, our mapping results (right) are strongly correlated with the official parade route (left), from Market/Beale to Market/8th Street in downtown San Francisco.
	
	
	
	\begin{figure}[t]
		\centering
		\includegraphics[width=1.0\linewidth,trim=0 0 0 0,clip]{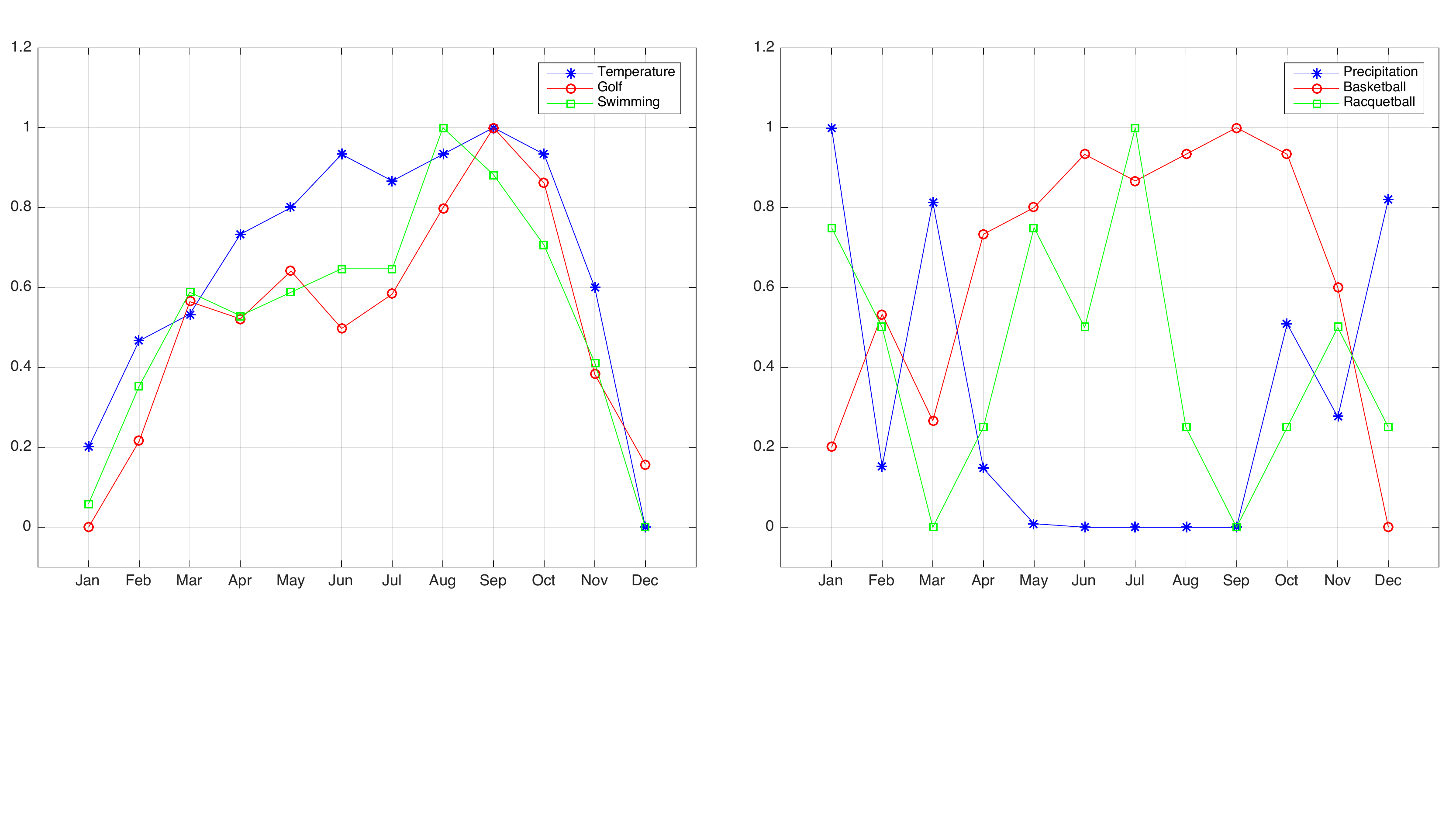}
		\vspace{-12ex}
		\caption{Weather impact on different activities. Left: Temperature. Right: Precipitation. }
		\label{fig:weather}
		\vspace{-2ex}
	\end{figure}
	
	\subsection{Weather Impact on Activities}
	\label{sec:weather_impact_on_human_activities}
	We here investigate if we can use our approach to observe the impact that weather has on activities. We separately consider temperature and precipitation\footnote{All weather data are obtained from https://www.wunderground.com/.}. We use the number of videos uploaded on a month-to-month basis to indicate the prevalence of particular activity. We plot this versus temperature or precipitation. All values are normalized from $0\sim1$ based on their maximum and minimum values.
	
	
	\noindent \textbf{Temperature}: As shown in Figure \ref{fig:weather} left, we observe a clear positive correlation between temperature and both golf and swimming. This makes sense because people are less likely to play golf or swim outdoors when it is cold.
	
	
	\noindent \textbf{Precipitation}: As shown in Figure \ref{fig:weather} right, precipitation has a great impact on outdoor activities like basketball. The curve for precipitation (blue) and basketball (red) are negatively correlated. By comparison, there is little correlation, positive or negative, for indoor activities such as racquetball.
	
	
	\begin{figure}[t]
		\centering
		\includegraphics[width=1.0\linewidth,trim=0 0 0 0,clip]{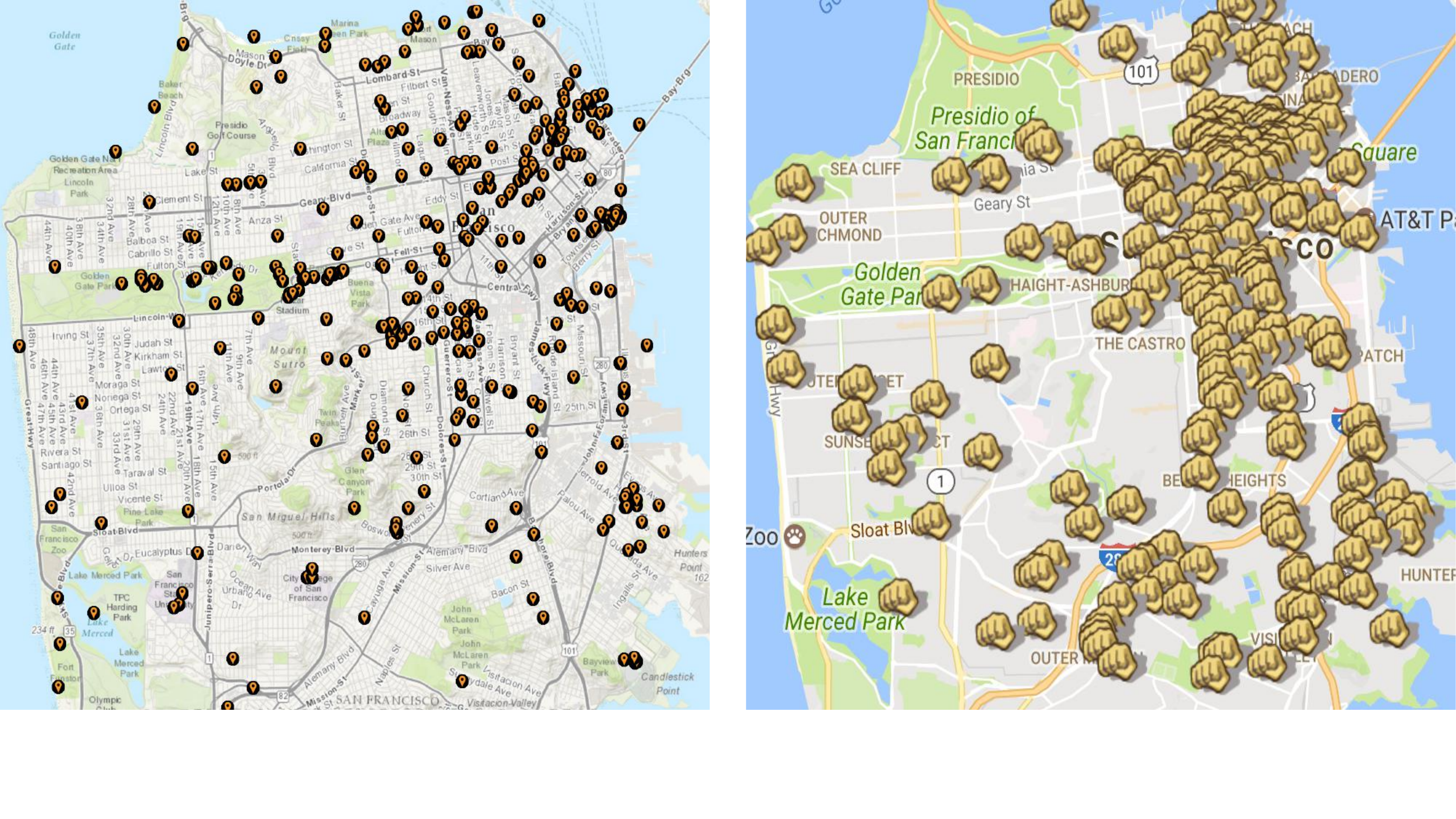}
		\vspace{-6ex}
		\caption{Violence detection. Left: our predicted street fight mapping. Right: the official police record of Assault mapping. }
		\label{fig:street_fight}
		\vspace{-2ex}
	\end{figure}
	
	\subsection{Crime Detection}
	\label{sec:violence_detection}
	
	Detecting criminal activities is important for public safety. We here demonstrate how our framework can be used to map violence using YouTube videos. This shows how our framework can generalize to a range of applications related to smart cities given suitable training data.
	
	We apply our framework to the San Francisco YouTube videos and detect $7,784$ instances of street fight. The locations of the videos are shown in Figure \ref{fig:street_fight} left. We notice concentrations of violence in downtown San Francisco, the Mission District, Hunters Point, etc. These are known to be high-crime areas. For comparison, we show the locations of Assault from a San Francisco crime map in Figure \ref{fig:street_fight} right\footnote{Data source is from: https://spotcrime.com/ca/san+francisco} derived from official police records. Our predicted locations are shown to be correlated with the official records.
	
	We would like to point out how our framework is different and complementary to using traditional surveillance cameras to monitor crime. We use geo-tagged videos from YouTube. The challenge is that these videos are not taken from the same viewpoint, with the same camera, with controlled lighting conditions, etc. This makes our problem much more difficult. However, we are able to leverage the scale and embedded perspective of the crowd to detect incidents that might not be captured using surveillance cameras.
	
	
	
	\begin{figure}[ht]
		\centering
		\subfigure[Tag/Title Based Method]{\includegraphics[width=0.45\linewidth,trim=0 0 0 0,clip]{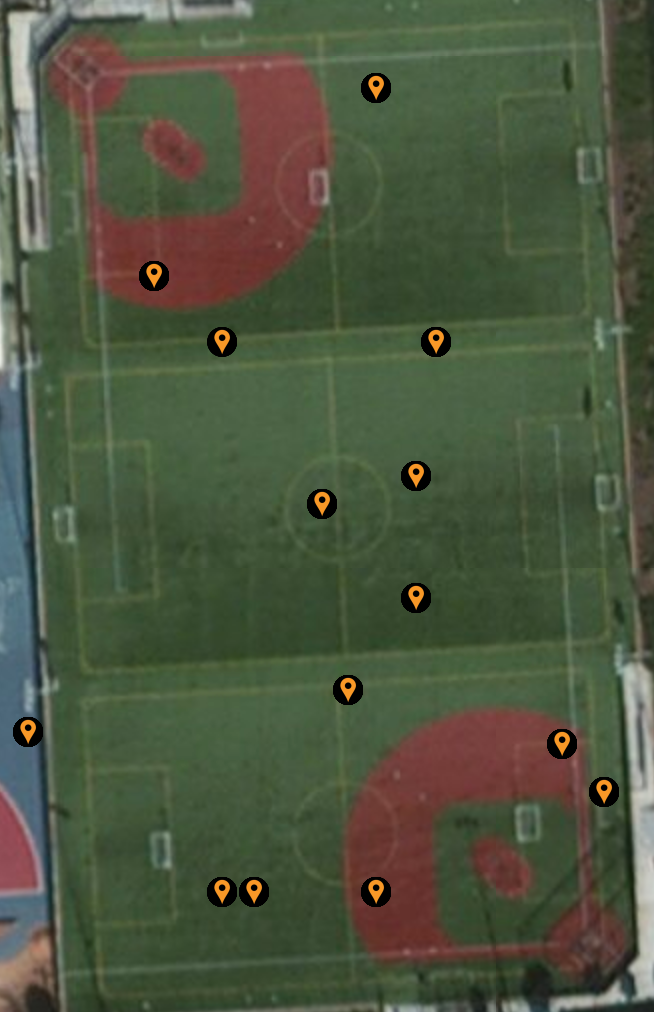}\label{fig:football_soccer_false}}\hspace{2pt}
		\subfigure[Content Based Method (ours)]{\includegraphics[width=0.45\linewidth,trim=0 0 0 0,clip]{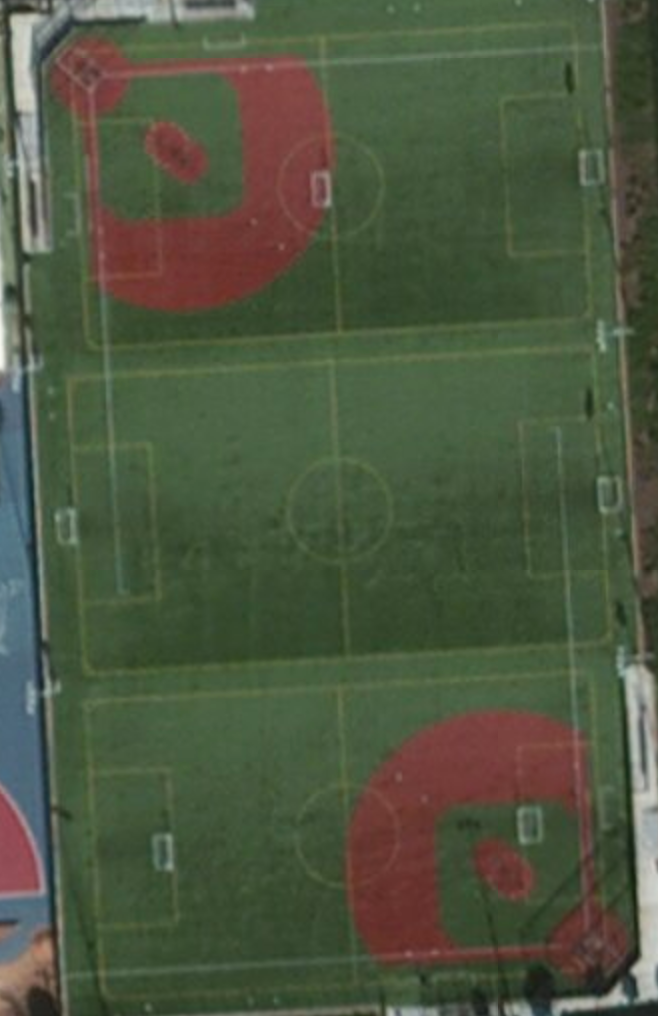}\label{fig:football_soccer}}\hspace{0pt}
		\vspace{-2ex}
		\caption{The advantage of using the visual content of videos instead of the tags/titles. The left image shows the locations of videos with ``football'' in the tag/title. These are actually soccer videos. The right image shows our football detector does not produce such false positives.}
		\label{fig:content_advantage}
	\end{figure}
	
	\section{Discussion}
	\label{sec:discussion}
	
	\noindent \textbf{Advantage over Tag/Title-Based Mapping}
	
	There are many advantages of using the visual content of geo-tagged videos instead of the tags/titles to map activities. Visual data is rich and can convey much more about what is occurring at a location than tags/titles. The visual content is also not subject to the ambiguity or imprecision of language. We demonstrate this here by mapping football and soccer using the San Francisco YouTube videos.  
	
	Most of the world uses football to refer to what is called soccer in the United States. This can cause problems when using tags/titles to map these two sports. Figure \ref{fig:content_advantage} shows a satellite image of several soccer fields at Ulloa elementary school in San Francisco's Sunset District. The image on the left shows the locations of videos with ``football'' in the tag/title such as ``football campaign'' and ``kids playing football''. Closer inspection shows these are not really videos of football but are of soccer. By comparison, the image on the right shows the results of our football activity detector. Using the visual content allows us to avoid the false positive detections in this case.

	\section{Conclusion}
	\label{sec:conclusion}
	We performed the first investigation into using the visual content of geo-tagged videos to map human activity. We utilized the recent hidden two-stream networks to detect 10 different activities in a large collection of YouTube videos of San Francisco. Our approach can run in real time which is important for real time applications. We performed a series of experiments to show our framework can map a diverse set of activities, can map specific events such as parades and street fights, can observe the impact that weather has on activities and is more accurate than using the tags/titles of the videos.
	
	In the future, we plan to investigate whether our framework can be adapted to detect a range of suspicious activities in surveillance video such as theft, vandalism, etc. A challenge to using CNNs for surveillance video is the relative lack of training data. We will explore whether YouTube videos can be used to at least pre-train the models. The challenge will then be to generalize the models to the different viewpoint, etc. of the surveillance videos.
	
	Additional directions include scaling the mapping to country or continental regions as well as to more activity classes. Finally, we will investigate reducing the sizes of our CNN models so they can be deployed in mobile devices or at endpoint equipment such as networked cameras.

	\section{Acknowledgments}
	We gratefully acknowledge the support of NVIDIA Corporation through the donation of the Titan X GPU used in this work. This work was funded in part by a National Science Foundation CAREER grant, \#IIS-1150115, and a seed grant from the Center for Information
	Technology in the Interest of Society (CITRIS).

	\bibliographystyle{ACM-Reference-Format}
	\bibliography{sigproc} 
	
\end{document}